\documentclass[10pt,twocolumn,letterpaper]{article}

\usepackage[pagenumbers]{cvpr}      %
\usepackage{duckuments}
\usepackage[usenames,dvipsnames]{xcolor}

\makeatletter
\DeclareRobustCommand\onedot{\futurelet\@let@token\@onedot}
\def\@onedot{\ifx\@let@token.\else.\null\fi\xspace}

\makeatother

\newcommand{\boldparagraph}[1]{\vspace{0.0cm}\noindent{\bf #1.}}

\definecolor{darkgreen}{rgb}{0,0.7,0}
\definecolor{darkyellow}{rgb}{0.8,0.8,0}
\definecolor{bittersweet}{rgb}{1.0, 0.44, 0.37}
\definecolor{amber}{rgb}{1.0, 0.49, 0.0}
\definecolor{lgray}{rgb}{0.83,0.83,0.83}

\definecolor{color_unlabled}{rgb}{0.0,0.0,0.0}
\definecolor{color_vehicle}{rgb}{0.0,0.0,0.56}
\definecolor{color_road}{rgb}{0.5,0.25,0.5}
\definecolor{color_redlight}{rgb}{1.0,0.0,0.0}
\definecolor{color_person}{rgb}{0.859,0.078,0.234}
\definecolor{color_roadline}{rgb}{0.613,0.914,0.195}
\definecolor{color_sidewalk}{rgb}{0.953,0.137,0.906}
\definecolor{teaser_red}{RGB}{222,112,97}

\definecolor{ellisred}{rgb}{0.87,0.44,0.38} %
\definecolor{ellisgreen}{rgb}{0.69,0.90,0.52} %
\definecolor{elliscyan}{rgb}{0.29,0.77,0.74} %
\definecolor{ellisorange}{rgb}{0.89,0.55,0.28} %
\definecolor{ellisblue}{rgb}{0.41,0.61,0.86} %

\definecolor{tuedgray}{RGB}{56,55,55}
\definecolor{tuelgray}{RGB}{246,246,246}
\definecolor{tuedblue}{RGB}{26,58,91}
\definecolor{tuelblue}{RGB}{133,203,210}
\definecolor{tueoblue}{RGB}{119,221,204}
\definecolor{tueogreen}{RGB}{119,221,159}
\definecolor{tuesgreen}{RGB}{186,213,72}
\definecolor{tueyellow}{RGB}{255,221,0}
\definecolor{tuered}{RGB}{234,75,46}

\mathchardef\mhyphen="2D

\definecolor{cvprblue}{rgb}{0.21,0.49,0.74}
\usepackage[pagebackref,breaklinks,colorlinks,citecolor=cvprblue]{hyperref}
\usepackage{multirow}
\usepackage{multicol}
\usepackage{makecell}
\usepackage{xcolor}
\usepackage{algorithm} 
\usepackage{amssymb}
\usepackage{amsfonts}
\usepackage{graphicx}
\usepackage{booktabs}
\usepackage{xr}
\usepackage{epigraph}
\usepackage[T1]{fontenc}
\definecolor{mycitecolor}{rgb}{0, 0.4, 0.7}
\usepackage{comment}
\usepackage{subcaption}
\usepackage{amsmath}
\usepackage{algorithm}
\usepackage{array}
\usepackage[switch]{lineno}
\usepackage{longtable}
\usepackage{xspace}
\usepackage{url}
\usepackage{overpic}
\usepackage{ragged2e}
\usepackage{xpatch}
\usepackage{pifont}
\usepackage{enumitem}
\usepackage{bbm}
\usepackage{colortbl}
\usepackage{soul}
\usepackage{adjustbox}
\usepackage{hhline}
\usepackage{algpseudocode}

\usepackage{stfloats}
\usepackage{xr-hyper}
\usepackage{hyperref}
\usepackage[font={small}]{caption} %

\usepackage{etoolbox}
\makeatletter
\pretocmd{\chapter}{\addtocontents{toc}{\protect\addvspace{15\p@}}}{}{}
\pretocmd{\section}{\addtocontents{toc}{\protect\addvspace{5\p@}}}{}{}
\pretocmd{\subsection}{\addtocontents{toc}{\protect\addvspace{3\p@}}}{}{}
\makeatother
\usepackage{titletoc}

\def\paperID{22} %
\def\confName{CVPR}
\def\confYear{2024}

\title{
CarLLaVA: Vision language models for camera-only closed-loop driving
}

\author{
Katrin Renz$^{1,2,3\ast}$ \quad
Long Chen$^{1}$ \quad
Ana-Maria Marcu$^{1}$ \quad
Jan Hünermann$^{1}$ \\
Benoit Hanotte$^{1}$ \quad
Alice Karnsund$^{1}$ \quad
Jamie Shotton$^{1}$ \quad
Elahe Arani$^{1}$ \quad
Oleg Sinavski$^{1}$ \\
[2mm]
$^1$~Wayve\quad
$^2$~University of Tübingen \quad
$^3$~Tübingen AI Center \quad
}

\begin{document}

\maketitle

{\let\thefootnote \relax \footnote{$^*$Work done while interning at Wayve.}}
{\let\thefootnote \relax \footnote{Project video: \href{https://youtu.be/E1nsEgcHRuc}{ \texttt{https://youtu.be/E1nsEgcHRuc}}}}

\vspace{-0.3cm}
\begin{abstract}
\vspace{-0.3cm}

In this technical report, we present CarLLaVA, a Vision Language Model (VLM) for autonomous driving, developed for the CARLA Autonomous Driving Challenge 2.0. CarLLaVA uses the vision encoder of the LLaVA VLM and the LLaMA architecture as backbone, achieving state-of-the-art closed-loop driving performance with only camera input and without the need for complex or expensive labels. Additionally, we show preliminary results on predicting language commentary alongside the driving output.
CarLLaVA uses a semi-disentangled output representation of both path predictions and waypoints, getting the advantages of the path for better lateral control and the waypoints for better longitudinal control.
We propose an efficient training recipe to train on large driving datasets without wasting compute on easy, trivial data. 
CarLLaVA ranks 1st place in the sensor track of the CARLA Autonomous Driving Challenge 2.0 outperforming the previous state-of-the-art by 458\% and the best concurrent submission by 32.6\%.

\vspace{-0.5cm}

\end{abstract}
\section{Introduction}
\label{sec:intro}

The trend in autonomous driving is shifting towards end-to-end solutions,
showed by recent advances in industry \cite{templeton2024tesla} and the state-of-the-art performance on the CARLA Leaderboard 1.0~\cite{jaeger2023hidden, shao2023reasonnet, shao2022safetyenhanced, wu2022trajectoryguided, chen2022learning}. 
Most of the top-performing entries on the CARLA Leaderboard 1.0~\cite{2020c} rely on expensive LiDAR sensors, with the exception of TCP \cite{wu2022trajectoryguided}, which employs a camera-only approach. Additionally, multi-task learning has emerged as a common strategy for enhancing performance~\cite{Chitta2023PAMI}. However, this requires access to labels, such as BEV semantics, depth, or semantic segmentation, which are expensive to obtain in the real world.
This makes it hard to transfer insights from research using simulators to real world driving in a scalable and cost-efficient way. CarLLaVA in contrast only relies on commonly available and easy to obtain driving data such as camera images and driving trajectory and is a camera only method. \\
Additionally, most state-of-the-art CARLA methods use ResNet-style backbones pretrained on ImageNet~\cite{jaeger2023hidden, shao2023reasonnet, shao2022safetyenhanced, wu2022trajectoryguided}. However, recent progress in pretraining techniques, such as CLIP~\cite{Radford2021ICML}, MAE \cite{He2021ARXIV}, and DINO, have demonstrated the advantages of using Vision Transformers (ViTs) \cite{Sharir2021ICLR} over traditional CNN-encoders for improved feature learning. Moreover, state-of-the-art VLMs~\cite{liu2023llava,li2023blip2,chen2024internvl} that fine-tune the CLIP encoder exhibit nuanced image understanding, indicating the existence of strong vision features. CarLLaVA makes use of this by using the vision encoder of LLaVA-NeXT~\cite{liu2023llava,liu2023improvedllava, liu2024llavanext} which is pre-trained on internet-scale vision-language data. While the size of modern VLMs could be viewed as a concern for inference time when deployed on real vehicles, several recent works showed that this is a solvable engineering problem~\cite{lingo2,nuro,wang2023driveanywhere}. \\
In this technical report, we describe the details of our driving model CarLLaVA, which includes the following properties and advantages:
\textbf{Camera only without expensive labels}: Our method only uses camera input, eliminating the need for additional expensive labels such as Bird’s Eye View (BEV), depth, or semantic segmentation. This label-free approach reduces dependency on extensive labeled datasets, making deployment on real cars more feasible.
\textbf{Vision-Language Pretraining}: Our approach leverages a vision encoder pre-trained on internet-scale vision-language data. We demonstrate that this pretraining can be effectively transferred to the task of driving, resulting in improved driving performance compared to training from scratch on driving data.
\textbf{High-resolution input}: We noticed that the default resolution of the CLIP vision encoder is not sufficient for quality driving. Similar to LLaVA\cite{liu2024llavanext}, we split input images into patches to allow the VLM access smaller details in the driving images such as distant traffic lights and pedestrians. In contrast to LLaVA we do not use the small resolution global patch to reduce the number of tokens.
\textbf{Efficient Training Recipe}: We propose an efficient training recipe that makes more use of interesting training samples, significantly reducing training time.
\textbf{Semi-Disentangled Output Representation}: We propose a semi-disentangled representation with both time-conditioned waypoints and space-conditioned path waypoints, leading to better control.

\section{Related Work}

\boldparagraph{Foundation models for driving} 
Recently, large language models (LLMs) have been integrated into driving systems to leverage their reasoning capabilities for addressing long-tail scenarios. Multi-modal LLM-based driving frameworks such as LLM-Driver \cite{chen2023driving}, DriveGPT4 \cite{xu2023drivegpt4}, and DriveLM \cite{sima2023drivelm} utilize foundation models with the inputs from different modalities for driving.
GPT-Driver \cite{mao2023gptdriver} and LanguageMPC \cite{sha2023languagempc} fine-tune ChatGPT as a motion planner using text. Knowledge-driven approaches \cite{wen2023dilu, fu2023drive} are also adopted to make decisions based on common-sense knowledge and evolve continuously. However, most of these works have been evaluated primarily through qualitative analysis or in open-loop settings.
The most similar works leveraging foundation models for closed-loop driving in CARLA are DriveMLM \cite{wang2023drivemlm} and LMDrive \cite{shao2023lmdrive}, which utilize multi-modal LLMs. However, these approaches rely on image and LiDAR inputs with customized encoders, without leveraging the power of vision-language pretraining and focused on tasks like instruction following. In comparison we focus on pure closed-loop driving performance to provide a baseline that can solve basic driving behaviors to enable future research on VLMs for driving.

\boldparagraph{End-to-end closed-loop driving in CARLA}
End-to-end training based on Imitation Learning (IL) is the dominant approach for
state-of-the-art methods on the CARLA Leaderboard 1.0~\cite{Chen2022CVPRa, Wu2022NeurIPS, Shao2022CORL, jaeger2023hidden}. Those methods are mostly incorporate numerous auxiliary outputs and rely on expensive sensors like LiDAR. In contrast, we build a model that only relies on camera images and the driving trajectory. \\
The dominant output representation is predicting waypoints with a GRU and using PID-controllers for lateral and longitudinal control~\cite{Chitta2022PAMI, Renz2022CORL, Chen2022CVPRa, Wu2022NeurIPS, Shao2022CORL, Shao2023CVPR, Jia2023CVPR, Zhang2023CVPR, jaeger2023hidden}. TCP~\cite{Wu2022NeurIPS} showed that waypoints perform poorly in turns, but predicting direct control performs worse in avoiding collisions. They propose a situation-based fusion strategy of those representations. Interfuser~\cite{Shao2022CORL} proposed predicting path waypoints together with a combination of forecasting and heuristics to obtain control. TF++~\cite{jaeger2023hidden} uses path waypoints for lateral control and target speed classes for longitudinal control. In our work we leverage the path representation for improved steering together with the standard waypoints for longitudinal control avoiding heuristics or the need for predefined classes. Additionally directly predict the waypoints from the output features of the transformer without using GRU. 
\section{Method}
\label{sec:method}

In the following sections, we provide a comprehensive overview of our architecture and training methodology.

\begin{figure}[t]
\begin{center}
   \includegraphics[width=\linewidth]{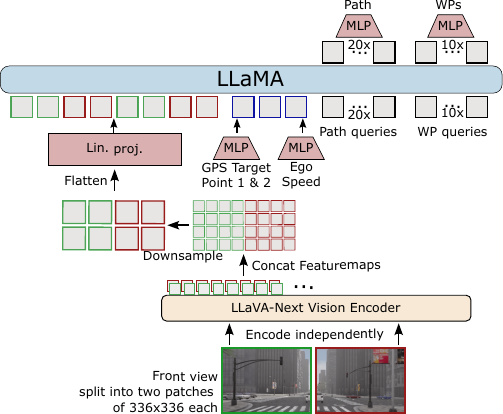}
\end{center}
\vspace{-0.4cm}
\caption{\textbf{CarLLaVA base model architecture. (C1T1)} The images are split in two, and each split is independently encoded and then concatenated, downsampled and projected into a pre-trained large language model. The output utilises a semi-disentangled representation with both time-conditioned waypoints and space-conditioned path waypoints for improved lateral control. }
\label{fig:architecture}
\end{figure}

\boldparagraph{Task}
The objective is to reach a specified target location on a 10x10 $km^{2}$ map while passing predetermined intermediate target points. The map includes diverse environments such as highways, urban streets, residential areas, and rural settings, all of which must be navigated under various weather conditions, including clear daylight, sunset, rain, fog, and nighttime scenarios.
Along the way the agent must manage various complex scenarios such as encountering pedestrians, navigating parking exits, executing unprotected turns, merging into ongoing traffic, passing construction sites or avoiding vehicles with opening doors.

\boldparagraph{Architecture} An overview of our base architecture can be seen in Fig.~\ref{fig:architecture}. \\
\textit{Input/Output Representation.} 
The model inputs include camera images, the next two target points, and the ego vehicle's speed. We tested several configurations: (1) the base model (C1T1) with a single front view image, (2) the temporal model (C1T2) which includes image features from the previous timestep, and (3) the multi-view model (C2T1) which adds a low-resolution rear-view camera to the high-resolution front view.
For the output, we use a semi-disentangled representation with both time-conditioned waypoints with a PID controller for longitudinal control and space-conditioned path waypoints with a PID controller for lateral control. Early experiments with entangled waypoints led to steering errors, especially during turns or when swerving around obstacles. By using path waypoints, we achieve denser supervision, as we also predict the path when the vehicle is stationary, leading to improved steering behaviour. For longitudinal control we use standard time-conditioned waypoints to make use of the better collision avoidance compared to directly predicting control~\cite{Wu2022NeurIPS}. We also experimented with target speed classification and GRUs, but these methods did not perform as well, although we lack official performance metrics.\\
\textit{HD-Vision Encoder.} To encode the camera images, we use the LLaVA-NeXT vision encoder, specifically the CLIPViT-L-336px model, which is the highest resolution trained CLIP model. High-resolution images are crucial for driving because important information, such as traffic lights at large intersections, may only be visible in a few pixels. To leverage CLIP pre-training at higher resolutions than 336x336, we use LLaVA's \textit{anyres} technique \cite{liu2024llavanext}. We divide high-resolution images into multiple large patches of up to 336x336 pixels, encoding each independently, and then concatenating the resulting features in spatial dimension to form a single large feature map for the original image.
Using a VLM not only provides strong features, but also offers the advantage of easily query the VLM to identify what information are captured in the image features.
More specifically, we queried the VLM for example for the state of traffic lights at different input resolutions to determine the optimal resolution and therefore the number of patches. \\
\textit{Adapter.} To reduce computation overhead due to the nature of the quadratic complexity of the LLaMA transformer, we downsample the feature map to half the number of tokens. After flattening, we employ a linear projection layer to map the vision features to the embedding space of the language model. To encode the target points and ego speed, we utilize a multi-layer perceptron (MLP) following a normalization layer. Additionally, we add camera encodings for the different views (model C2T1) and temporal encodings when using images from multiple time steps (only for model C1T2).\\
\textit{LM-Decoder.} We use the LLaMA architecture as a decoder. In addition to the sensor input tokens, we use learnable queries to generate the path and waypoints. An MLP on top of the output features generates waypoint differences. The cumulative sum of these differences yields the final waypoints, which are supervised during training using mean squared error (MSE) loss.
For our preliminary results on generating language explanations we auto-regressively sample the language explanation after generating the path and waypoints. During training we feed the tokenized explanations and use a standard language modelling (LM) loss. We use the tokenizer and LM-head of the pretrained Tiny-LLaMA model. \\

\begin{table}
\footnotesize
\centering
    \setlength{\tabcolsep}{3pt}
    \begin{tabular}{l l| l l| c c c } %
         & \textbf{Method} & \textbf{Sensors} & \textbf{Aux. Labels}& \textbf{DS} $\uparrow$ & \textbf{RC} $\uparrow$ & \textbf{IS} $\uparrow$ \\ %
        \hline
        \parbox[t]{2mm}{\multirow{4}{*}{\rotatebox[origin=c]{90}{Map}}}
        &CaRINA modular	& L+C+M & OD & 1.14 &	3.65 &	0.46 	\\ %
        &greatone & undisclosed & undisclosed &	2.17	&10.78	& 0.37 \\
        &Kyber-E2E &  L+C+R+M & IS, OD &	3.47	&8.48	&\textbf{0.50} \\ %
         &TF++ & L+C & SS, D, OD, BS &\textbf{5.56}&	\textbf{11.82}	&0.47 \\
        \hline
        \hline
        \parbox[t]{2mm}{\multirow{4}{*}{\rotatebox[origin=c]{90}{Sensor}}}
        &CARLA & priv. & priv. &	0.25 &	15.20 &	0.10 \\ %
        &Zero-shot TF++	& L+C & SS, D, OD, BS &0.58 &	8.53 &	0.38 \\ %
        & CaRINA hybrid & L+C & IS, OD &	1.23	& 9.56	& 0.31 \\	%
        &TF++ & L+C & SS, D, OD, BS &5.18	&11.34	&\textbf{0.48}	\\
        \hline
        &CarLLaVA (ours) & C & - & \textbf{6.87} & \textbf{18.08} & 0.42 \\ %
        \hline
    \end{tabular}
    \vspace{-0.2cm}
    \caption{\textbf{Leaderboard 2.0 Results.} CarLLaVA achives state-of-the-art performance on the leaderboard. Legend: L: Lidar, C: Camera, R: Radar, M: Map, priv: privileged, OD: Object Detection (3D position and pose), IS: Instant Segmentation, SS: Semantic Segmentation, D: Depth, BS: BEV semantics.}
    \label{tab:leaderboard}
    \vspace{-0.2cm}
\end{table}

\begin{table*}[t]
\vspace{-0.5cm}
\centering
\footnotesize
\subfloat[
\textbf{Output.}
\label{tab:output}
]{
\centering
\begin{minipage}{0.26\linewidth}{\begin{center}
\begin{tabular}{l|r r}
\toprule
& \multicolumn{1}{c}{\textbf{DS} $\uparrow$} & \textbf{Stat} $\downarrow$\\
\midrule
WPs    & 3.21 &0.68\\
+Path     & \cellcolor{lgray} 4.49 &0.0 \\
\bottomrule
\end{tabular}
\end{center}}\end{minipage}
}
\hspace{0.6em}
\subfloat[
\textbf{Vision encoder}.
\label{tab:encoder}
]{
\begin{minipage}{0.2\linewidth}{\begin{center}
\begin{tabular}{l|r}
\toprule
\textbf{}
& \multicolumn{1}{c}{\textbf{DS} $\uparrow$}\\
\midrule
LLaVA & \cellcolor{lgray} 6.87  \\ %
- pretraining & 0.45 \\
Resnet-34 &  2.71 \\
\bottomrule
\end{tabular}
\end{center}}\end{minipage}
}
\hspace{0.6em}
\subfloat[
\textbf{Early stopping}.
\label{tab:early}
]{
\begin{minipage}{0.2\linewidth}{\begin{center}
\begin{tabular}{l|r}
\toprule
\textbf{}
& \multicolumn{1}{c}{\textbf{DS} $\uparrow$}\\
\midrule
1300  &  3.93\\
1800 & 4.49 \\
2100 & \cellcolor{lgray} 6.87 \\
2400 & 6.35 \\
\bottomrule
\end{tabular}
\end{center}}\end{minipage}
}
\hspace{0.6em}
\vspace{-.7em}
\caption{Ablations of different parts of our model, showcasing the superiority of the semi-disentangled output representation and the large impact of the correct threshold for early stopping. The score of the default configuration is highlighted in gray. All numbers are official Leaderboard scores.}
\label{tab:ablations} \vspace{-2em}
\end{table*}

\boldparagraph{Efficient training of large models}
Our models have between 350M and 1.3B parameter. To finetune these large models on our task we rely on models pretrained on internet-scale data, a large dataset and an efficient training recipe which is described in the following. \\
\textit{Dataset}
We utilize the privileged rule-based expert \textit{PDM-light} \cite{Beißwenger2024TECH} to collect a dataset. We divide the official CARLA routes of Town 12 and Town 13 into shorter segments centered around scenarios to reduce trivial data (e.g., driving straight without any hazardous events) and simplify data management. We use short routes with a single scenario as proposed by\cite{Chitta2023PAMI, li2024think2drive}, however with the introduction of Leaderboard 2.0, the maximum distance between target points increased from 50 meters to 200 meters. The short routes often fall within this distance, causing a distribution shift, as the next target point is the end of the route (i.e, closer than 200m) rather than the position that would be used when having long routes. Consequently, we employ a second set of routes featuring three scenarios per route.
To ensure balance, we adjust the number of routes per scenario and apply random weather augmentation and modify the parameter \textit{distance} for scenarios by ±10\%. Overall, we collect 2.9 million samples at 5 fps. \\
For the language generation experiment we use the logic of the rule-based expert to generate explanations. More precisely, we use the leading object obtained from the experts' Intelligent Driver Model (IDM)\cite{Treiber_2000} as well as information about changing the path to swerve around objects. In addition, we use heuristics based on the ego waypoints to distinguish between driving intentions like starting from stop or keep driving at the same speed. As this experiment is only intended to showcase the potential of using LLMs for driving, we do not add detailed statistics of the obtained labels and keep it for future work. \\
\textit{Buckets.}
The majority of driving involves straight, uneventful segments. To maximize the collection of interesting scenarios during data collection, we focus on capturing a diverse range of challenging situations. However, some ratio of easy and uneventful data is inevitable. Training models on the entire dataset revealed that straight driving without hazards is effectively learned in the early epochs, resulting in wasted compute in later epochs as the models continue to train on these uninteresting samples. To address this issue, we create data buckets containing only the interesting samples and sample from these buckets during training instead of the entire dataset. 
We use: (1) five buckets for different amount of acceleration and deceleration with one specifically for starting from stop, excluding samples with acceleration between -1 and 1, (2) two buckets for steering, excluding samples for going straight, (3) three buckets for vehicle hazard with vehicles coming from different directions, (4) one for stop sign, red light and walker hazards each, (5) one bucket for swerving around obstacles and (6) one bucket that samples from the whole dataset to keep a small portion of uneventful data such as driving straight.
This approach reduces the number of samples per epoch to 650,000.

\section{Experiments}
\label{sec:experiments}

\boldparagraph{Benchmarks} 
\textit{Leaderboard2.0.} We use the official test server with secret routes under different weather conditions.
\textit{10xShort.} For the models where we could not get Leaderboard results, we use a local evaluation on short routes with one scenario per route to evaluate the models ability to solve each scenario type. We use maximum 10 routes per scenario which are randomly sampled from the whole set. 

\boldparagraph{Metrics}
We report the official CARLA metrics, Driving Score (DS), Route Completion (RC) and Infraction Score (IS).
DS is calculated in a way that the reduction due to infractions does not linearly correlate with the increase in DS due to higher RC (i.e., with a constant infraction per km the DS gets much worse for higher RC for models that can solve the scenarios below a certain percentage). Forcing the agent to stop a route early can maximize DS.

\boldparagraph{Implementation Details}
We use a learning rate of 3e-5 with a cosine annealing schedule. The batch size of our base model is 20, while for specific configurations, we use a batch size of 10 for C1T2 and a batch size of 12 for C2T1. The AdamW optimizer is employed with a weight decay of 0.1.
Our vision encoder consists of 305 million parameters. We experiment with the LLaMA architecture in three configurations: LLaMA-50M, LLaMA-350M (both trained from scratch), and a 1B TinyLLaMA with LoRA finetuning~\cite{hu2021lora}, applied to all linear layers as demonstrated to be effective by QLoRA~\cite{dettmers2024qlora}.
We apply the same data augmentation techniques as TF++~\cite{jaeger2023hidden} but with more aggressive shift and rotation augmentation (shift: 1.5m, rot: 20 deg). Additionally, we add histogram enhancements to improve the contrast and quality of input images for night time driving. DeepSpeed v2 is utilized for optimizing training efficiency and memory usage. We train for 30 epochs. Our base model, C1T1, trains in approximately 27 hours using 8xA100 40GB GPUs.
During inference we apply early stopping to counter the nature of DS described in the metric section. We track the travelled distance and stop driving after a specified distance when the steering angle is close to zero to prevent stopping in the middle of an intersection where it could happen that other vehicles crash into us.

\boldparagraph{Results} \\
\textit{Leaderboard state of the art.}
We present the official Leaderboard results in Tab.~\ref{tab:leaderboard}. With our base-model CarLLaVA C1T1 we outperfrom the state of the art (5.18 vs 6.87 DS). However, we observed a high variance on the Leaderboard score, detailed results on mean and standard deviation can be found in the supplementary (the official Leaderboard numbers are our first submissions of the models, the repetitions to calculate mean and std happened after the challenge deadline).
It is also noteworthy that, to the best of our knowledge, our model is the only model on the leaderboard working only with camera images and without the usage of additional auxiliary labels (note: for the new entry \textit{greatone} we do not know what their method is). \\
\textit{Output representation.} Tab.~\ref{tab:output} compares the DS on the Leaderboard for the different output representations. As the goal of the additional path prediction is improved lateral control, we also report the collisions with static layout as this is mainly caused due to bad steering. With the semi-disentangled representation we can reduce the layout collision from 0.68 to 0.0 showcasing the strength of additional path predictions. \\
\textit{Vision-Language and CLIP pretraining.} We ablate the pretraining of the vision encoder and train the same model from scratch. Tab.~\ref{tab:encoder} '-pretraining' shows that the pretraining stage is essential for good driving performance (more tuning of the training hyperparameters can further improve the performance but is unlikely to reach the performance of the pretrained model). Additionally, we show a comparison to the widely used Resnet-34 pretrained on ImageNet. The decreased performance (2.71 vs. 6.87 DS) indicates the importance of the larger ViT and the internet-scale image-language pretraining. \\
\textit{Early stopping.}
We ablate the thresholds for the early stopping as it is not trivial to calculate the perfect trade-off as the routes and density of scenarios are secret (however a rough function of the expected DS can be caluculated which we used to get a rough idea). Tab.~\ref{tab:early} shows the Leaderboard DS for a given travelled distance in meters. This hyperparameter has a big impact on the final score.

\begin{figure}[t]
    \begin{center}
       \includegraphics[width=\linewidth]{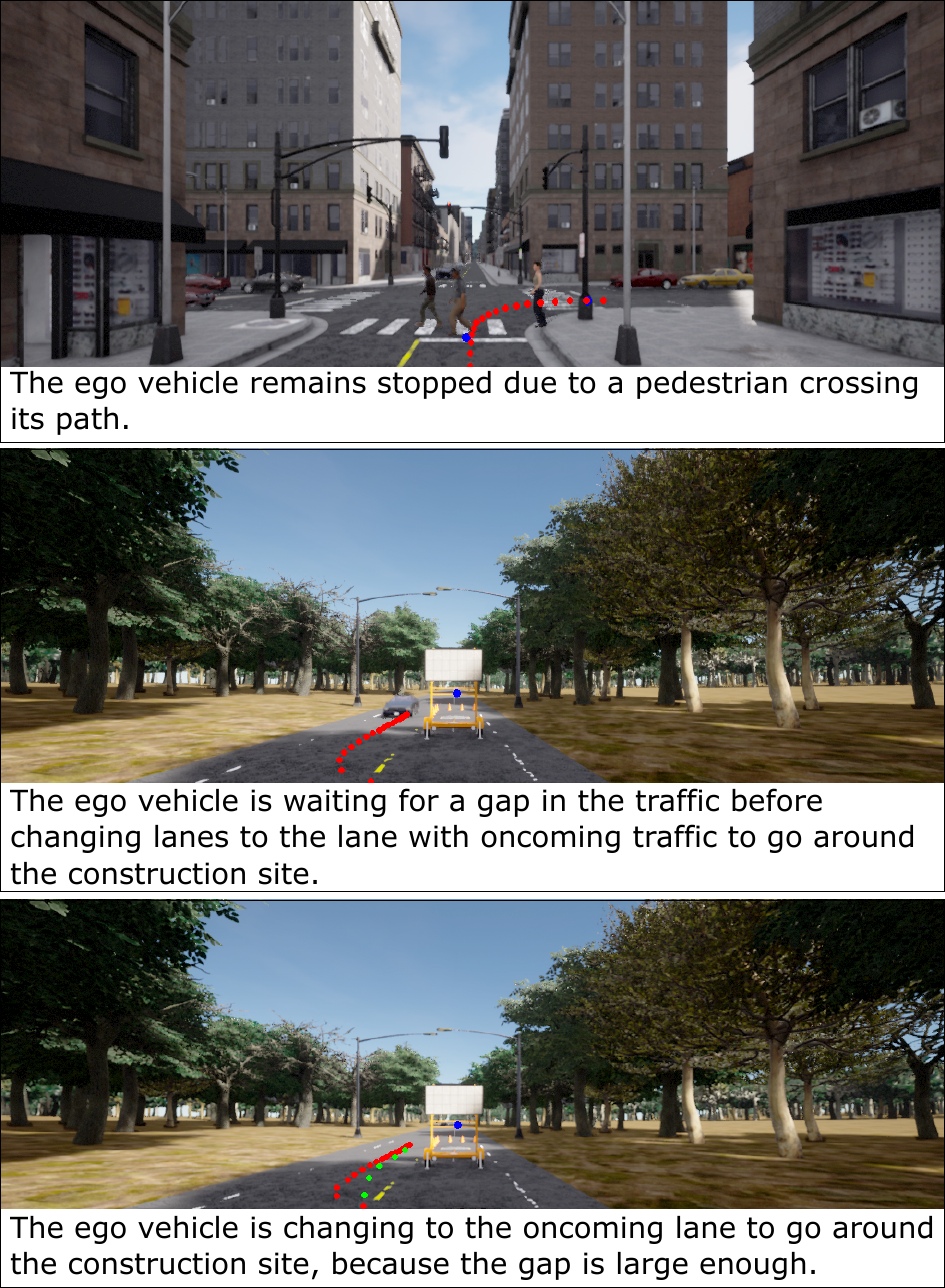}
    \end{center}
    \vspace{-0.4cm}
    \caption{\textbf{Qualitative examples of generated language.} Red: predicted path, Green: predicted waypoints, Blue: Target Points}
    \label{fig:qual}
    \vspace{-0.3cm}
\end{figure}

\boldparagraph{Preliminary Results} \\
\begin{table}[t]
\footnotesize
\subfloat[
\textbf{Scale}.
\label{tab:scale}
]{
\begin{minipage}{0.45\linewidth}{\begin{center}
\begin{tabular}{l|r}
\toprule
\textbf{}
& \multicolumn{1}{c}{\textbf{DS\textsubscript{S}} $\uparrow$}\\
\midrule
50M & \cellcolor{lgray} 90.40  \\ %
 350M  &  92.49  \\ %
1B pt LoRA &  90.03 \\
1B s LoRA &  89.57 \\
\bottomrule
\end{tabular}
\end{center}}\end{minipage}
}
\hspace{0.6em}
\subfloat[
\textbf{Input}. %
\label{tab:input}
]{
\begin{minipage}{0.45\linewidth}{\begin{center}
\begin{tabular}{l|r}
\toprule
\textbf{}
& \multicolumn{1}{c}{\textbf{DS\textsubscript{S}} $\uparrow$}\\
\midrule
default   & \cellcolor{lgray} 90.40 \\
+ temporal    & 90.37  \\ %
+ back    & 88.81 \\ %
- pretraining & 75.43 \\
\bottomrule
\end{tabular}
\end{center}}\end{minipage}
}
\caption{Further ablations of different parts of our model. The score of the default configuration is highlighted in gray. DS\textsubscript{S} is performance on the \textit{10xShort} benchmark.}
\vspace{-0.3cm}
\label{tab:prelablations}
\end{table}
\label{app:sec:ablation}
In addition to our ablations we show preliminary results to showcase the potential to extend to multiple views and temporal input, scaling our base model and adding language predictions.\\
\textit{Leaderboard variance.} We submitted our base model CarLLaVA C1T1 with an early stopping threshold of 2100 and 2400 three times to the leadboard to get an estimate of the evlauation variance. For the 2100 model we obtain the following scores: 5.5, 6.8 and 5.3 resulting in a mean DS of 5.87 with a standard deviation of 0.81. The base model with a threshold of 2400 obtained 6.3, 6.3 and 4.8 resuting in a mean of 5.8 with standard deviation of 0.87. \\
\textit{Scale.} In an additional experiment we scale up the LLaMA architecture (Tab.~\ref{tab:scale}). 
 Training a 350M parameter model from scratch improves performance slightly. However scaling to 1B parameter and finetuning with LoRA resulted in worse performance for using a pretrained LLM (pt) and training from scratch (s). We suspect that this may be due to the use of LoRA finetuning and not fully tuned hyperparameters, but further investigation is needed. This remains an interesting research question for future work. \\
 \textit{Extending the input.} To be able to fully solve autonomous driving, information from more than one camera (especially for camera-only arhcitectures) and temporal information are needed. In Tab.~\ref{tab:input} we show results for a model with temporal information and one with added back camera. Qualitative investigations showed improvements in the expected scenarios (less rear-end collisions for \textit{+temporal} and improved lane-change behaviour for \textit{+back}). Interestingly the overall score does not increase.\\
\textit{Language explanations.} With the additional language training our model is able to produce commentary that comments the current driving behaviour (Fig.~\ref{fig:qual}). This is not intended as an actual explanation as the training misses an important grounding step (i.e., commentary is not always aligned with the actions the model takes). We leave this for future work.

\boldparagraph{Failure cases} The most common failure cases of our model are rear end collision, which can be reduced by using the temporal input of the C1T2 model and maneuver like merging especially in high speeds.
\section{Conclusion}
\label{sec:discussion}
In this report, we present CarLLaVA the winning entry in the CARLA Autonomous Driving Challenge 2.0 2024, which leverages vision-language pretraining and uses only camera images as input. By utilizing a semi-disentangled output representation and an efficient training approach, CarLLaVA demonstrates superior performance in both lateral and longitudinal control. Its ability to operate without expensive labels or sensors makes it a scalable and cost-effective solution. The results indicate a substantial improvement over previous methods, showcasing the potential of vision-language models in real-world autonomous driving applications.

\vspace{0.2cm}
\boldparagraph{Acknowledgements} 
We thank Kashyap Chitta, Julian Zimmerlin, Jens Beißwenger, Bernhard Jäger and Andreas Geiger for valuable discussions and help with the expert. We also thank the International Max Planck Research School for Intelligent Systems (IMPRS-IS) for supporting K. Renz.

{
    \small
    \bibliographystyle{ieeenat_fullname}
    \bibliography{bibliography_short, bibliography_custom, bibliography}
}

\clearpage

\end{document}